\documentclass[sigconf]{acmart}

 \AtBeginDocument{  \providecommand\BibTeX{{    \normalfont B\kern-0.5em{\scshape i\kern-0.25em b}\kern-0.8em\TeX}}}

\usepackage{Definitions}
\usepackage{epstopdf}
\usepackage{subfig} 
\usepackage[export]{adjustbox}
\usepackage{booktabs}

\usepackage{xspace}

\newcommand{\dscold}{\textsc{Dsc-Exact}\xspace}
\newcommand{\dscour}{\textsc{Dsc-Approx}\xspace}
\newcommand{\ambold}{\textsc{Amb-Exact}\xspace}
\newcommand{\ambour}{\textsc{Amb-Approx}\xspace}

\setcopyright{rightsretained}
\copyrightyear{2021}
\acmYear{2021}
\acmConference[AIES '21]{Proceedings of the 2021 AAAI/ACM Conference on AI, Ethics, and Society}{May 19--21, 2021}{Virtual Event, USA}
\acmBooktitle{Proceedings of the 2021 AAAI/ACM Conference on AI, Ethics, and Society (AIES '21), May 19--21, 2021, Virtual Event, USA\\}
\acmDOI{10.1145/3461702.3462630}
\acmISBN{978-1-4503-8473-5/21/05}

\begin{document}
\fancyhead{}

\title[Accounting for Model Uncertainty in Algorithmic Discrimination]{Accounting for Model Uncertainty in \\Algorithmic Discrimination}
\author{Junaid Ali}
\email{junaid@mpi-sws.org}
\affiliation{  \institution{Max Planck Institute for \\Software Systems}
        \country{Germany}
  }

\author{Preethi Lahoti}
\email{plahoti@mpi-inf.mpg.de}

\affiliation{  \institution{Max Planck Institute for \\Informatics}
      \country{Germany}}

\author{Krishna P. Gummadi}
\email{gummadi@mpi-sws.org}
\affiliation{  \institution{Max Planck Institute for \\Software Systems}
    \country{Germany}
}

\renewcommand{\shortauthors}{Ali et al.}

\begin{abstract} 
 
Traditional approaches to ensure group fairness in algorithmic decision making aim to equalize ``total'' error rates for different subgroups in the population. In contrast, we argue that the fairness approaches should instead focus only on equalizing errors arising due to \emph{model uncertainty} (a.k.a epistemic uncertainty), caused due to lack of knowledge about the best model or due to lack of data. In other words, our proposal calls for ignoring the errors that occur due to uncertainty inherent in the \emph{data}, i.e., aleatoric uncertainty. We draw a connection between \emph{predictive multiplicity} and \emph{model uncertainty} and argue that the techniques from predictive multiplicity could be used to identify errors made due to model uncertainty. 
We propose scalable convex proxies to come up with classifiers that exhibit predictive multiplicity and empirically show that our methods are comparable in performance and up to four orders of magnitude faster than the current state-of-the-art. We further propose methods to achieve our goal of equalizing group error rates arising due to model uncertainty in algorithmic decision making and demonstrate the effectiveness of these methods using synthetic and real-world datasets.

\end{abstract}



\maketitle

\section{Introduction} 

Prediction systems are being used for several socially impactful tasks, e.g., predicting recidivism risk in order to help judges make bail decisions, assessing credit ratings, assessing the risk of defaulting on a loan and predicting the risk of accident for insurance purposes. This development has raised concerns about prediction systems being discriminatory. To address this concern, researchers have proposed a class of group fairness methods, which seek to equalize overall errors across different groups 
of sensitive attributes such as gender or race \cite{zafar_preferred,hardt_nips16,ali2019,zafar_fairness}. This approach treats all errors as equal. 
However, not all errors are the same. 

\begin{figure}[tb!]
	\centering
	\includegraphics[angle=0, width=0.85\columnwidth]{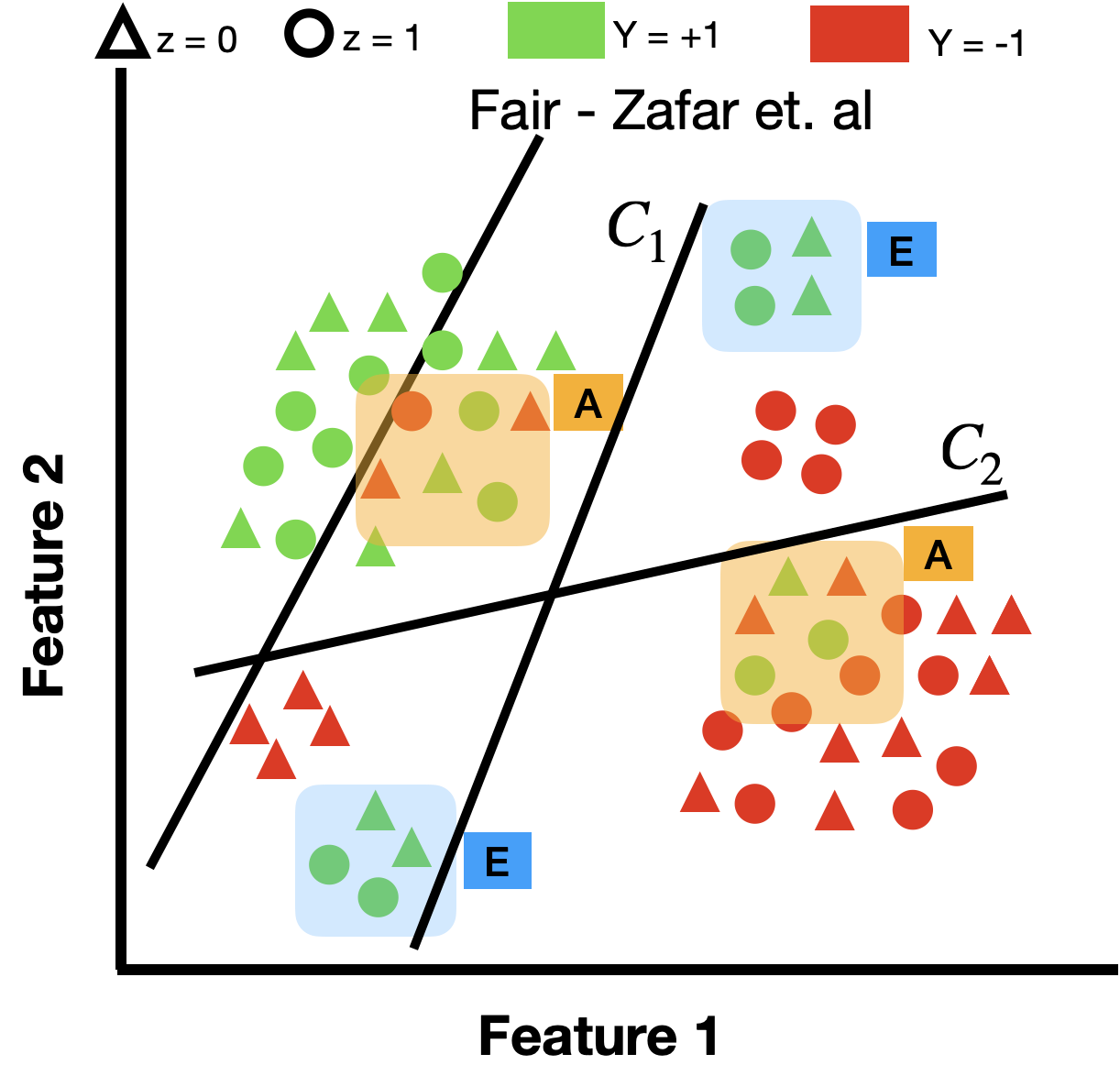}
	\caption{Illustrative example: 
		Consider a binary classification task with two features and a sensitive feature (z) represented by the shape of the points, i.e., circles and triangles. 
				Green and red colors represent ground truth positive and negative labels, respectively. Classifiers $C_1$ and $C_2$ are equally accurate classifiers achieving $80\%$ accuracy. The difference between false positives of triangles and circles for $C_1$ is $32\%$ and $-25\%$ with $C_2$. However, these two classifiers disagree on their decision on $29\%$ of the data, i.e., which lies in the ambiguous region between the two classifiers. The errors made by these classifiers in the ambiguous regions marked by \textbf{E} are \emph{epistemic errors}. While the errors highlighted by the region \textbf{A} are \emph{aleatoric errors}. If we were to pick one of these classifiers as the final decision boundary it would be unfair to the points receiving a favorable decision with the other classifier. On the other hand, a fair classifier equalizing false positive rates, using~\cite{zafar_fairness}, gives an accuracy of only $68\%$. However, as it does not disregard the aleatoric uncertainty it changes the decisions of several points that clearly belong to the positive cluster.	 	
	}
	\label{fig:motivating_example}
\end{figure}

It is well-known that errors in prediction models arise out of both epistemic (model) uncertainty and aleatoric (inherent) uncertainty \cite{hora1996aleatory,depeweg2018decomposition, malinin2019uncertainty}. Equalizing total error could lead to unjustifiably wrong decisions for some datapoints. Consider Figure~\ref{fig:motivating_example}, where a traditional 
fair classifier that equalizes total errors including the irreducible ones that arise due to aleatoric uncertainty. This results in many datapoints getting a negative outcome even though they clearly belong to the positive cluster. These errors are particularly consequential in socially impactful applications. 

In this paper, we argue to distinguish between the errors caused by different types of uncertainty. Specifically, we introduce the notions of \emph{aleatoric errors} and \emph{epistemic errors}. We refer to the errors that occur only due to model or epistemic uncertainty as \emph{epistemic errors} and the ones that occur due to aleatoric uncertainty, we call the \emph{aleatoric errors}. Figure~\ref{fig:motivating_example} shows an example of both types of errors. The errors made by the classifiers $C_1$ and $C_2$ that are highlighted by the region \textbf{A} are due to the noise in the data, as these wrongly predicted datapoints are surrounded by predominantly the other class label, i.e., ground truth positive or ground truth negative datapoints. We refer to these types of errors as \emph{aleatoric errors}. While the errors in the region marked by \textbf{E} are due to model uncertainty as one could resolve this uncertainty by gathering more data or by choosing a more complex model. These types of errors are \emph{epistemic errors}. Our proposal is to \emph{ignore} the aleatoric errors which are likely to be irreducible due to inherent uncertainty in the data or the prediction task at hand and we argue to \emph{only} equalize the epistemic errors, i.e., the ones that occur due to methodological limitations. 

In order to identify the epistemic errors that are caused by model uncertainty, we leverage the work on predictive multiplicity by \citet{marx2019predictive}.  \emph{Predictive multiplicity} refers to the scenario where multiple {predictive models} have similar predictive performance (e.g., similarly accurate) but assign contradictory predictions on a subset of the datapoints, which characterize the \textit{ambiguous regions}. We draw a connection between predictive multiplicity and \textit{model uncertainty}. 

Model uncertainty is defined as the level of spread or 'disagreement' in the decisions of an ensemble sampled from the posterior~\cite{malinin2019uncertainty}. We use predictive multiplicity to identify model uncertainty, i.e., we argue that the disagreement in equally well performing models signals uncertainty in the model parameters. Specifically, we argue that if the classifiers exhibiting predictive multiplicity are chosen from a complex enough hypothesis class, then the regions in the feature space with high model uncertainty that are likely to have the epistemic errors would coincide with the ambiguous regions produced by predictive multiplicity. Therefore, our proposal of equalizing only the epistemic errors translates into equalizing errors in the ambiguous regions, while ignoring the ones in the unambiguous regions.

One of the \emph{key properties} of our proposal is that people whose outcomes are affected by our fairness requirements are the people whose outcomes are ambiguous or uncertain in the first place. Put differently, we do not alter the outcomes of people with unambiguously positive or negative outcomes. In contrast, current methods for achieving equal error rates might alter outcomes for people with unambiguous outcomes as well, creating a difficult accuracy-fairness tradeoff dilemma. We believe that our proposal would be easier to justify in many practical scenarios.

Key technical contributions of our approach are (a) designing efficient and scalable methods for identifying ambiguous regions, and (b) designing mechanisms for equalizing group error rates in the ambiguous regions. In order to solve the first challenge, we propose \emph{convex proxies} to find models that exhibit predictive multiplicity. For the second challenge, our key insight is \textit{to reuse the highly accurate models trained to identify the ambiguous regions} in the first place. Specifically, given the set of classifiers identifying ambiguous regions, we propose to \emph{stochastically pick a classifier} from this set when making a decision. 
The probabilities of picking the classifiers are chosen in a way that equalizes group error rates in the ambiguous regions in expectation. 
An additional benefit of our approach compared to the traditional way of making a deterministic decision is that we account for model uncertainty by introducing stochasticity in our predictions, and thus many datapoints in the \emph{ambiguous region} have a non-zero probability of receiving a favorable outcome. As there is some chance of getting a favorable outcome for most datapoints affected by our fairness notion, it would make our proposal more desirable than the traditional approach of assigning decisions deterministically. \\

\noindent \textbf{Contributions and Outline:} 
\begin{itemize}
	\item \textbf{Conceptual contribution:} We argue that uncertainty in prediction should be accounted for when designing fairness approaches. To this end, we propose to \emph{only} equalize errors occurring due to model uncertainty, i.e., the epistemic errors. 
	\item \textbf{Technical contributions:} i) We propose tractable scalable convex proxies to identify \emph{ambiguous regions}. That is, for a given dataset $\Dcal$, we identify a set of linear or nonlinear classifiers that are equally accurate, but they conflict in their predictions for a subset of datapoints (see Section \ref{sec:scal_pred_mult}). ii) We also formalize our proposal to only equalize the epistemic errors and present a fairness approach that equalizes group errors in the \emph{ambiguous regions} (see Section \ref{sec:fairness_amb}).
	
	\item \textbf{Empirical contributions:} i) Our experimental results show that our proposed scalable convex proxies to identify regions with predictive multiplicity are comparable in performance and up to four orders of magnitude faster than the current state-of-the-art (see Section \ref{sec:eval_synth}, Table~\ref{tab:table_comp}). ii) Our experimental results on a synthetic and two real-world datasets show that our fairness methods improve fairness in the ambiguous regions while achieving comparable accuracy to the best classifier (see Sections \ref{sec:eval_synth} and \ref{sec:eval_real}).

\end{itemize}

\section{Preliminaries and Background}
In this section, we present the necessary background on binary classification and predictive multiplicity. 
\subsection{Binary Classification}
 Given a training 
 dataset $\mathcal{D} = \{ (\xb_i, y_i) \}_{i=1}^{N}$, the goal of a binary classifier is to learn a function $\phib: \RR^{d} \to \{-1,1\}$ between the feature
  vectors $\xb \in \RR^d$ and the class labels $y \in \{-1, 1\}$. In order to learn this function one has to solve $\phib^* = \argmin_{\phib} \Rb_{\Dcal}(\phib): \Rb_\Dcal(\phib) =  \frac{1}{N} \sum_{\xb_i,y_i} \II{[\phib(\xb_i) \neq y_i]}$. However, this function is non-convex in $\phib$ and worse, it is intractable, which makes it especially difficult to solve for large datasets. In the rest of the text we drop the subscript, $\Dcal$, for brevity. To efficiently solve the problem, it is a standard practice to use a convex proxy. One minimizes a given convex loss $L(\thetab)$ over $\Dcal$, \ie, $\thetab^{*} = \argmin_{\thetab} L(\thetab)$, in order to find $\thetab^{*}$ for convex decision-boundary-based classifiers like linear/non-linear SVM and logistic regression, where $\thetab \in \RR^{d}$.
Then, for a given (potentially unseen) feature vector $\xb$, 
one predicts the class label 
$\hat{y} = 1$ if $d_{\thetab^{*}}(\xb) \geq 0$ and 
$\hat{y} = -1$ 
otherwise, where $d_{\thetab^{*}}(\xb)$ denotes the signed distance from 
$\xb$ to the decision boundary. For convenience, we define $\thetab^{*}(\xb) = 1$ if $d_{\thetab^{*}}(\xb) \geq 0$ and $\thetab^{*}(\xb) = -1$ if $d_{\thetab^{*}}(\xb) < 0$.

In the rest of the paper, we consider $\thetab_{best}$ to be the most accurate classifier yielded by minimizing logistic regression loss with L2 regularizer, where weights of the regularizer were picked based on the performance on the validation set. Similarly, we consider $\phib_{best}$ to be the best classifier using 0-1 loss ($\Rb_\Dcal$), selected using a validation set.

\subsection{Background on Predictive Multiplicity}
In this section, we formally introduce the notion of predictive multiplicity and discuss the existing measures and mechanisms to compute predictive multiplicity. 

\xhdr{Predictive multiplicity} A prediction problem exhibits predictive multiplicity if one can find a classifier $\phib$ for a given small value $\epsilonb$ such that $\Rb(\phib) - \Rb(\phib_{best}) <= \epsilonb$, and there exists at least one datapoint with feature vector $\xb_i$ such that $\phib(\xb_i) \neq \phib_{best}(\xb_i)$ \cite{marx2019predictive}. The definition for classifiers trained with proxy loses is similar. One could consider $\epsilonb$  to be $0$ but in practice a classifier that is slightly less accurate on the training data might be equally or even more accurate on the test data. 

Predictive multiplicity is defined for a set of two or more classifiers, referred to as the $\epsilonb$-level set. Given the most accurate classifier $\phib_{best}$, the $\epsilonb$-level set of $\phib_{best}$ is a set of classifiers which have an accuracy only up to $\epsilonb$ lower than $\phib_{best}$. Formally, over the dataset $\Dcal$, $ \CC_{\epsilonb,\phib_{best}} = \{\phib: \Rb(\phib) - \Rb(\phib_{best}) \leq \epsilonb\}$.

\xhdr{Measures of predictive multiplicity}
\citet{marx2019predictive} propose two measures for predictive multiplicity for a given set of classifiers, namely \emph{Discrepancy} and \emph{Ambiguity}. 

For a given set of classifiers, \emph{Discrepancy} is defined as the maximum fraction of the datapoints on which any classifier in the set disagrees on the outcomes with the most accurate classifier. Formally, given $\CC_{\epsilonb,\phib_{best}}$ and dataset $\Dcal$, 

\begin{equation}\label{eq:non_scal_disc}
 \delta_{\epsilonb}(\phib) = \max_{\phib \in \CC_{\epsilonb}} \frac{1}{n} \sum_{\xb_i \in \Dcal} \II[\phib(\xb_i) \neq \phib_{best}(\xb_i)], 
\end{equation}
i.e., discrepancy is the maximum fraction of conflicting  decisions yielded by any classifier in $\CC_{\epsilonb,\phib_{best}}$ compared to $\phib_{best}$. 

\emph{Ambiguity} of a set of classifiers for a prediction task is defined as the fraction of datapoints given a different decision than the best classifier. Formally, given set $\CC_{\epsilonb,\phib_{best}}$ and dataset $\Dcal$, 
 \begin{equation}\label{eq:non_scal_amb}
  \alpha_{\epsilonb}(\phib) =  \frac{1}{n} \sum_{\xb_i} \max_{\phib \in \CC_{\epsilonb,\phib_{best}}} \II[\phib(\xb_i) \neq \phib_{best}(\xb_i)], 
 \end{equation}
 where $\max_{\phib \in \CC_{\epsilonb,\phib_{best}}} \II[\phib(\xb_i) \neq \phib_{best}(\xb_i)]$ is $1$ if there exists at least one classifier in $\CC_{\epsilonb,\phib_{best}}$ which gives a datapoint with features $\xb_i$ a different outcome than $\phib_{best}$, otherwise it is $0$. Hence, ambiguity is the fraction of datapoints on which any classifiers in $\CC_{\epsilonb,\phib_{best}}$ disagrees on the outcome with $\phi_{best}$. 

\xhdr{Methods to identify predictive multiplicity}
Inspired by the measures discrepancy and ambiguity, \citet{marx2019predictive} propose two methods that maximize these measures in order to find the classifiers that exhibit maximum predictive multiplicity for the given allowance of accuracy reduction. This would indicate the extent of predictive multiplicity for the prediction task at hand.

\xhdr{Exact discrepancy maximization (Dsc-Exact)}
Given a value of $\epsilonb$, the authors propose to train classifiers that minimize the agreement to $\phib_{best}$ under the constraint that its accuracy is only up to $\epsilonb$ lower than $\phib_{best}$, i.e., 
\begin{align}\label{eq:non_scal_disc_prob}
&\underbrace{\minimize_{\phib} \quad \sum_{\xb_i} \II[\phib(\xb_i) = \phib_{best}]}_\text{maximize discrepancy} \,\tag{P1}  \\
 &\mbox{subject to} \quad \underbrace{\Rb(\phib) \leq \Rb(\phib_{best}) + \etab}_\text{bound accuracy reduction} \nonumber
\end{align}
where $\etab \in (0,\epsilonb)$. One can obtain a set $\CC_{\epsilonb, \phib_{best}}$ by solving the above formulation for several $\etab$ values.

\xhdr{Exact ambiguity maximization (Amb-Exact)}
In order to find the classifiers that maximize the ambiguity measure for a given threshold of accuracy reduction, \citet{marx2019predictive} propose to train a classifier for each datatpoint in the training data that gives the datapoint a different decision than the most accurate classifier. Then, they pick the classifiers whose accuracy lies within the threshold of the allowed accuracy reduction. Specifically, 
they propose to train classifiers that change their decisions compared to $\phib_{best}$ for individual datapoints while minimizing 0-1 loss, i.e., 

 \begin{align}\label{eq:non_scal_amb_prob}
\underbrace{\minimize_{\phib}  \, \Rb(\phib)}_\text{maximize accuracy} \quad \text{subject to } \underbrace{\phib(\xb_i) \neq \phib_{best}(\xb_i)}_\text{change decision of $\xb_i$ w.r.t $\phib_{best}$} \, \forall \xb_i. \tag{P2}
\end{align}

Then, one can select $\CC_{\epsilonb,\phib_{best}}$ by pruning the set of classifiers resulting from the solution of the problem above, i.e., by selecting classifiers which are only $\epsilonb$ lower in accuracy than $\phib_{best}$. 

To solve both Problems \ref{eq:non_scal_disc_prob} and \ref{eq:non_scal_amb_prob}, \citet{marx2019predictive} propose mixed integer programming formulations. However, these formulations i) work only for linear classifiers and ii) have slow performance as these are exact, intractable and non-convex.

\section{Proposed approach}
\begin{figure*}[t]
 \centering

            \subfloat     {
     \includegraphics[angle=0, width=0.45\columnwidth]{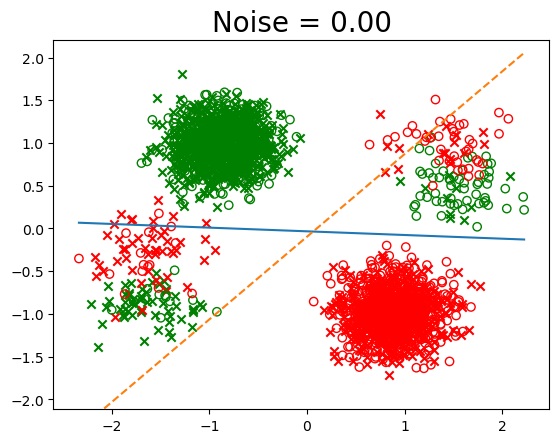}
         }
     \subfloat     {
     \includegraphics[angle=0, width=0.45\columnwidth]{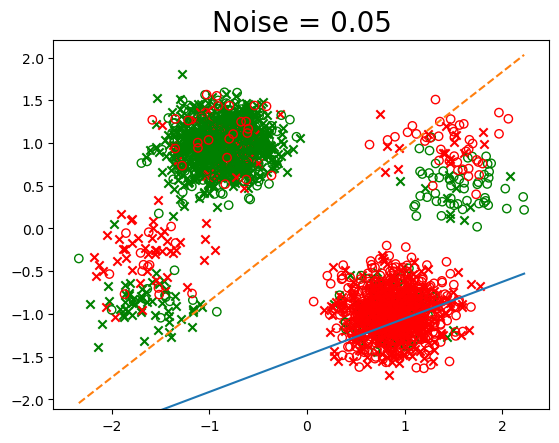}
         }
     \subfloat     {
         \includegraphics[angle=0, width=0.45\columnwidth]{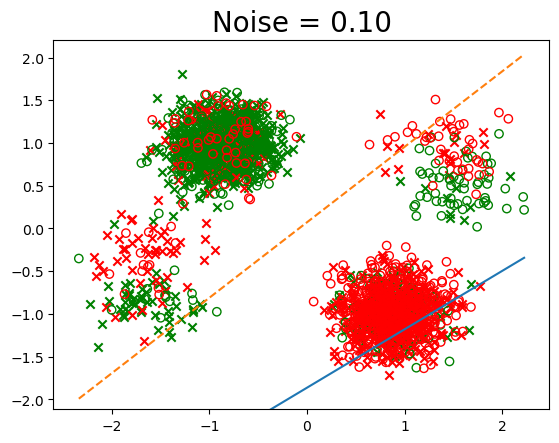}
         }
                             \subfloat     {
         \includegraphics[angle=0, width=0.65\columnwidth]{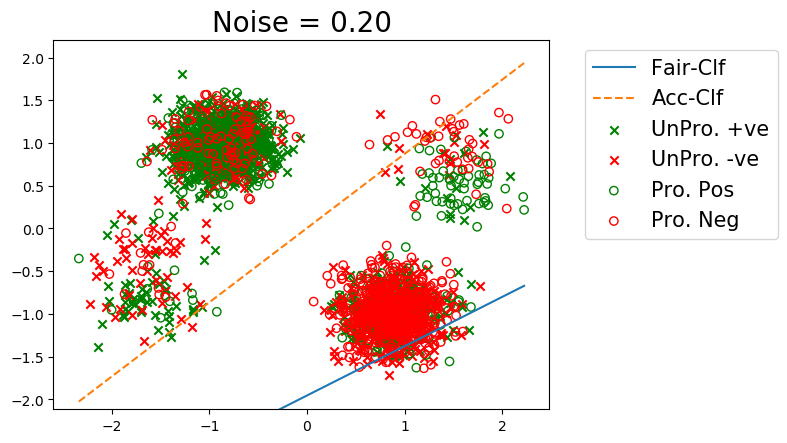}
         }
                    \caption{
    [Synthetic dataset] Figure demonstrates that state of the art fairness methods are effected by label noise.   
    }
         \label{fig:clf_fair}
\end{figure*}

\begin{figure*}[t]
 \centering
      \subfloat         {
     \includegraphics[angle=0, width=0.47\columnwidth]{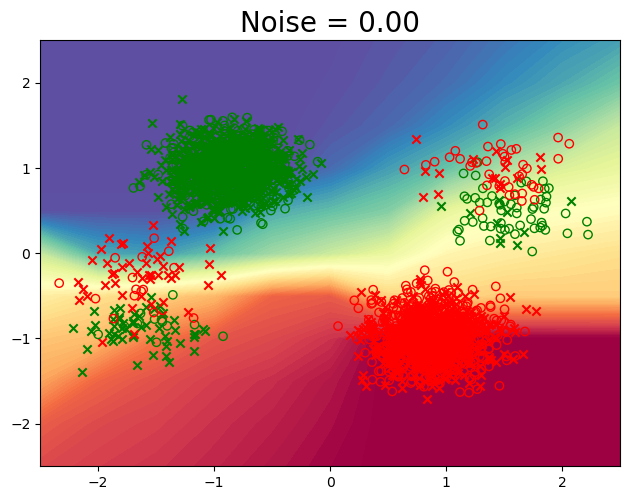}
         } 
     \subfloat     {
     \includegraphics[angle=0, width=0.47\columnwidth]{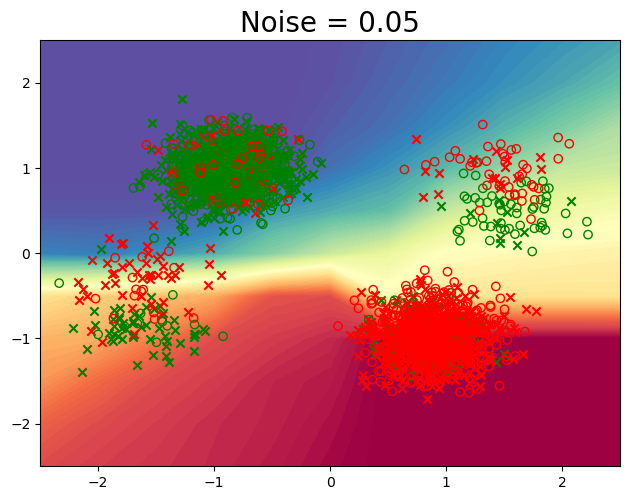}
         }
     \subfloat     {
         \includegraphics[angle=0, width=0.47\columnwidth]{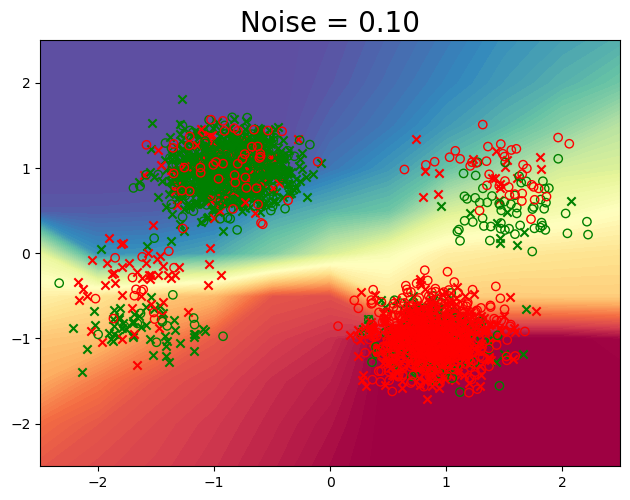}
         }
                              \subfloat     {
         \includegraphics[angle=0, width=0.47\columnwidth]{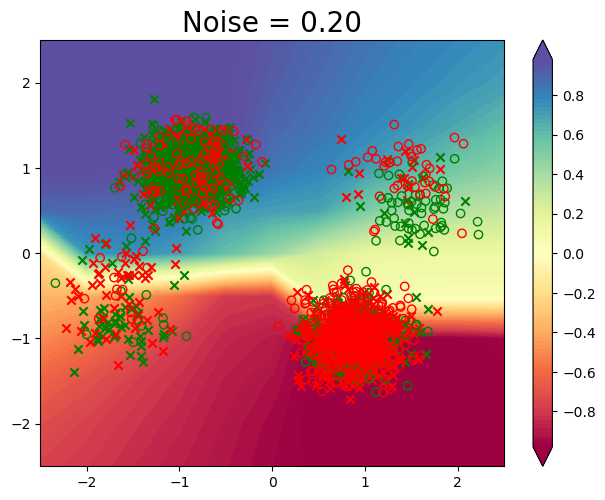}
         }
     
     \caption{
    [Synthetic dataset] Figure shows the expected class while equalizing FPRs using the classifiers solving \ref{eq:scal_amb_prob}. It demonstrates that our method is stable under label noise, as it consistently identifies same regions as ambiguous for different levels of noise values.  
    }

    \label{fig:clf_ours}
     \end{figure*}

In this section, 
we aim to answer the question: \textit{What is a fair model under model uncertainty}?  

We characterize model uncertainty using predictive multiplicity. 
Given a set of classifiers $\CC_{\epsilonb, \thetab_{best}}$ that exhibit predictive multiplicity, we consider $x_i$ to have an ambiguous decision if \textit{any} of the classifiers in $\CC_{\epsilonb,\thetab_{best}}$ gives it a conflicting decision compared to any other classifier. Formally a set of ambiguous points are defined as:  
\begin{equation}\label{eq:amb_region} 
\Acal := \{x_i: \thetab_{j}(x_i) \neq \thetab_k(x_i) \, \forall \, \thetab_j, \thetab_k \in \CC_{\epsilonb,\thetab_{best}} \}. \nonumber 
\end{equation}
These points characterize the ambiguous region. By choosing a single model from $\CC_{\epsilonb, \thetab_{best}}$ as the final model we might be unfair to some group in the ambiguous region. Our proposal of only equalizing the epistemic errors boils down to equalizing group error rates in the ambiguous region $\Acal$.

The \emph{key assumption} we make is that the hypothesis class for the classifiers, $\CC_{\epsilonb, \thetab_{best}}$, exhibiting  predictive multiplicity is sufficiently complex, i.e., if the data is nonlinearly separable the hypothesis class should include nonlinear classifiers. Under this assumption, all the errors in the the unambiguous region, i.e., where all the classifiers in the set $\CC_{\epsilonb, \thetab_{best}}$ agree in their decisions, would \emph{only} be due aleatoric uncertainty. The argument is as follows: Given the classifiers in set $\CC_{\epsilonb, \thetab_{best}}$ are picked from a sufficiently complex hypothesis class for the given data. Under this assumption, if all the classifiers agree in their prediction for a subset of the datapoints, then the resulting errors for these datapoints could only be due to inherent stochasticity of the prediction task or random noise, i.e., aleatoric errors. On the other hand, the ambiguous region, $\Acal$, would identify regions with high model uncertainty. The intuition is as follows: Given that the classifiers for set $\CC_{\epsilonb, \thetab_{best}}$ are chosen from a sufficiently complex hypothesis class. Under this assumption, if these equally accurate classifiers disagree on some datapoints this would include all the datapoints whose decisions are uncertain due to lack of data. This implies that \emph{all} the epistemic errors will lie in the ambiguous region. The ambiguous region could also have random noise hence causing some aleatoric errors. The results using the Synthetic dataset in Section~\ref{sec:eval_synth} confirm our hypotheses.

Next, we present our proposals for identifying the ambiguous region using scalable convex methods. Then, we discuss our methods for equalizing groups errors in the ambiguous region $\Acal$. 

\subsection{Scalable Methods for Predictive Multiplicity}\label{sec:scal_pred_mult}

In this section, we propose two convex methods to find the ambiguous region $\Acal$.

\xhdr{Approximate Discrepancy maximization (Dsc-Approx)} We propose the following convex and tractable proxy constraint that bounds similarity between $\thetab$ and $\thetab_{best}$, akin to the objective in Problem~\ref{eq:non_scal_disc_prob} that maximizes discrepancy: 
\begin{equation}\label{eq:scal_disc}
\frac{1}{N} \sum_x \max(0, d_{\thetab(x)} d_{\thetab_{best}(x)}) \leq \gamma, 
\end{equation}
where $d_{\thetab(x)}$ represents the distance of the datapoint with feature vector $\xb$ from the decision boundary of $\thetab$. $\max(0,\cdot)$ represents the agreement of decisions between $\thetab$ and $\thetab_{best}$. Specifically, if the decision for a subject with feature vector $\xb$ stays the same under $\thetab$ compared to $\thetab_{best}$, only then does the term $\max(0,\cdot)$ produce a non-zero number. Thus, by bounding the left hand side we are limiting the average allowed distance of the datapoints which have the same decisions under $\thetab$ and $\thetab_{best}$. Making this bound tighter would preferably admit $\thetab$ whose decisions are different on some of the datapoints than $\thetab_{best}$, as those datapoints contribute $0$ to the sum on the left hand side. This implies that one can control the number of decisions allowed to be the same between $\thetab$ and $\thetab_{best}$ by changing the value of $\gamma \in \RR+$. For example, $\gamma = +\infty$ would yield $\thetab = \thetab_{best}$ meaning that all the decisions between $\thetab$ and $\thetab_{best}$ are the same, i.e., $\thetab$ would yield a discrepancy of $0$ compared to $\thetab_{best}$. Similarly, for $\gamma = 0$ one aims to learn $\thetab$ whose decisions are different on all datapoints than to $\thetab_{best}$, i.e. a classifier yielding maximum discrepancy compared to $\thetab_{best}$. The value of $\gamma$ also controls the reduction in accuracy under $\thetab$ compared to $\thetab_{best}$. 

For linear boundary-based classifiers (logistic regression, linear SVM), $d_{\thetab}(\xb) = \thetab^T\xb$. For nonlinear SVM, one can write $d_{\beta}(\xb) = \sum_{i=1}^{N} \beta_i y_i k(\xb_i, \xb)$ for the optimization variables $\beta$ and a positive semidefinite kernel function $k(.,.)$. Hence, in both linear and nonlinear cases the constraint stays convex since the distance from the decision boundary is linear with respect to the optimization variables.

One can write a convex and tractable version of Problem \ref{eq:non_scal_disc_prob} using the logistic regression loss as follows: 

\begin{align}\label{eq:scal_disc_prob}
&\underbrace{\minimize_{\thetab} \quad - \frac{1}{N} \sum_{\xb_i,y_i} p(y_i|\xb_i; \thetab)}_\text{maximize accuracy} \tag{P3}\\
&\mbox{subject to} \quad \underbrace{\frac{1}{N} \sum_{\xb_i} \max(0, d_{\thetab(\xb_i)} d_{\thetab_{best}(\xb_i)}) \leq \gamma}_\text{enforce discrepancy} \nonumber
\end{align}
where $p(y = 1 | \xb, \thetab) = \frac{1}{1 + \exp(-\thetab^T\xb)}$. 

One can learn an appropriate $\gamma$ value using a validation set, for a given $\etab$ in \ref{eq:non_scal_disc_prob}. We construct $\CC_{\epsilonb,\thetab_{best}}$ by training classifiers with varying values of $\gamma$ and then picking the ones whose accuracy is only $\epsilonb$ lower than $\thetab_{best}$.

\xhdr{Approximate ambiguity maximization (Amb-Approx)} We propose the following convex and tractable constraints equivalent to the constraint in Problem \ref{eq:non_scal_amb_prob}. 
\begin{align}\label{eq:scal_amb}
d_{\thetab(\xb_i)} &< 0 \text{ if }  d_{\thetab_{best}(\xb_i)} \geq 0 \quad \forall \xb_i \\
d_{\thetab(\xb_i)} &\geq 0 \text{ if }  d_{\thetab_{best}(\xb_i)} < 0 \quad \forall \xb_i, \nonumber
\end{align}
where $d_{\thetab}$ is the distance from decisions boundary of $\thetab$. The constraints above require $\thetab$ to make a different decision than $\thetab_{best}$ on the datapoint $\xb_i$ . The constraints stay convex for both linear and nonlinear boundary based classifiers because one can write the distance from the decision boundary as a linear function of the optimization parameter in both cases. One can write a convex and scalable version of Problem~\ref{eq:non_scal_amb_prob} as follows: 
\begin{align}\label{eq:scal_amb_prob}
&\underbrace{\minimize_{\thetab} \quad - \frac{1}{N} \sum_{\xb_i,y_i} p(y_i|\xb_i; \thetab)}_\text{maximize accuracy} \tag{P4} \\
&\mbox{subject to }\quad d_{\thetab(\xb_i)} < 0 \text{ if }  d_{\thetab_{best}(\xb_i)} \geq 0 \quad \forall \xb_i  \nonumber \\ 
& \qquad \qquad \quad \underbrace{d_{\thetab(\xb_i)} \geq 0 \text{ if }  d_{\thetab_{best}(\xb_i)} < 0 \quad \forall \xb_i}_\text{change decision of $\xb_i$ w.r.t $\thetab_{best}$}, \nonumber 
\end{align}
where $p(y = 1 | \xb, \thetab) = \frac{1}{1 + \exp(-\thetab^T\xb)}$. We pick $\CC_{\epsilonb,\thetab_{best}}$ by training a set of classifiers which assign conflicting decisions to all the datapoints in the training set. Then, we pick the classifiers which are only $\epsilonb$ lower in accuracy than $\thetab_{best}$.

\subsection{Leveraging Predictive Multiplicity towards Fairness under Model Uncertainty}\label{sec:fairness_amb}
In this section, we propose to learn a meta classifier in order to equalize group errors 
arising 
due to model uncertainty. 

In order to do that, our key insight is to use the highly accurate classifiers that we trained to identify the ambiguous regions in the first place. Specifically, given the validation set of datapoints and $\CC_{\epsilonb,\thetab_{best}}$, picked by solving \dscour, \ref{eq:scal_disc_prob}, or \ambour, \ref{eq:scal_amb_prob}, we first identify the points with ambiguous decisions. We then construct a meta classifier by picking the classifiers stochastically from the set $\CC_{\epsilonb,\thetab_{best}}$. The probabilities for picking these classifiers are chosen in a way that aims to equalize group error rates on the ambiguous datapoints among different groups of a sensitive feature such as race or gender. For a binary valued sensitive feature $z = \{0,1\}$, we propose

\begin{align}\label{eq:amb_fairness}
&\minimize_{w} \quad |\sum_{\theta \in {\CC_{\epsilon}}} w_\theta \cdot \underbrace{({Err}_{z=1}(\theta) - {Err}_{z=0}(\theta)}_\text{FPR/FNR difference})| \tag{P5} \\
&\mbox{subject to} \quad 0 \le w_\theta \le 1 \quad \text{and} \quad  \sum_\theta w_\theta  = 1, \nonumber 
\end{align}
where ${Err}_{z=0}(\theta)$ and ${Err}_{z=1}(\theta)$ are false positive rates (FPR) or false negative rates (FNR) for group $0$ and $1$ of the sensitive feature \textit{in the ambiguous region}, $\Acal$. As the set of classifiers is predetermined, the error rates can be precomputed. Hence, the problem is convex and efficiently solvable, as the objective function is a linear function of optimization variable $w$.
 
The intuition is that the difference of the errors rates between the two groups, i.e., ${Err}_{z=1}(\theta) - {Err}_{z=0}(\theta)$, might be positive for some of the classifiers in $\CC_{\epsilonb,\thetab_{best}}$ and it might be negative for the others. We can then assign the probabilities $w_\theta$ to these classifiers in a way such that they cancel each others biases and the expected unfairness is minimized. Our experimental results on the real-world and synthetic datasets confirm our intuition (Tables~\ref{tab:synth_fairness},~\ref{tab:compas_fairness},~\ref{tab:sqf_fairness}). 

In the case of a non-binary valued sensitive feature, one can replace the error rate difference between two groups with pair-wise differences among all the groups. We learn the probability mass function $w$ using the validation datapoints, and when classifying the unseen test datapoints we use $w$ to pick the classifiers from $\CC_{\epsilonb, \thetab_{best}}$.

\section{Experiments}

In this section, we demonstrate the effectiveness of our methods using synthetic and real-word  datasets. Specifically, we answer the following evaluation questions:\\
\textbf{-- Q1.} How effective and fast are our methods in identifying the ambiguous regions? \\
\textbf{-- Q2.} What is the fairness and accuracy trade-off of our methods? \\
\textbf{-- Q3.} Are our methods robust to noisy data? \\

\subsection{Datasets}\label{sec:datasets}

We use a \textit{Synthetic dataset} because  i) we could easily alter the size of the datasets, which is useful as \dscold and \ambold have slow performance on larger complex datasets, especially with continuous valued features; ii) we could provide intuition for the type of ambiguous regions identified by our methods; iii) we could introduce noise in the data and check the robustness of our methods vs the existing methods. The data comprises $10000$ datapoints and $2$ features and a binary valued sensitive feature, $z$. The data is sampled from the following Gaussian distributions:
\begin{small}
\begin{align}
&\Ncal_1([-35; 65], [60, 1; 1, 120])\nonumber, \, \, 
\Ncal_2([15;-25], [60, 1; 1, 120])\nonumber, \\
&\Ncal_3([30; 65], [70, 1; 1, 100])\nonumber, \, \, \,
\Ncal_4([35;40], [70, 1; 1, 100])\nonumber, \\
&\Ncal_5([-55;5], [70, 1; 1, 100])\nonumber \, \text{and} \,
\Ncal_6([-55;-20], [70, 1; 1, 100])\nonumber \\\nonumber
\end{align}
\end{small}
From $\Ncal_1$, $4500$ points were sampled. Amongst these, $95\%$ of which were labeled ground truth positive and $65\%$ of these points were uniformly at random assigned to the non-protected class of the sensitive feature, i.e, $z=0$. A total of $4500$ points were sampled from $\Ncal_2$, $95\%$ of which are ground truth negative points and $65\%$ of these points were uniformly at random assigned to the protected class of the sensitive feature, i.e., $z=0$. Finally, $250$ points were sampled from $\Ncal_3$ and $\Ncal_5$ each, with ground truth negative labels, and $250$ points were sampled from $\Ncal_4$ and $\Ncal_6$ each and were assigned ground truth positive labels. $80\%$ of the points sampled from $\Ncal_3$ and $\Ncal_4$ and $20\%$ of the points sampled from $\Ncal_5$ and $\Ncal_6$, were uniformly at randomly assigned $z=1$. After sampling these points they were normalized to have a unit mean and a unit variance. A visual representation is shown in Figure~\ref{fig:clf_fair}. We flipped the class label of a fraction of datapoints which induced aleatoric errors through out the data. However, model uncertainty only exists in the sparse clusters shown in Figure~\ref{fig:clf_fair} as that could be reduced by gathering more data. Our hope is that predictive multiplicity would be able to identify regions with predominantly model uncertainty, i.e., the sparse clusters as the ambiguous regions for different levels of aleatoric uncertainty. We also experimented with other variations of the parameters and got similar results.

We processed the ProPublica \textit{COMPAS dataset}~\cite{propublica_data} similar to \citet{zafar_dmt}, which resulted in $5,287$ subjects and $7$ features. Given these features we have to predict whether a criminal defendant would recidivate within two years (positive class) or not (negative class). We consider race, with  values African-Americans, $z=0$, and white, $z=1$, to be a sensitive feature in this dataset.

The NYPD \textit{SQF dataset} comprises features of pedestrians, such as race, gender, height \etc~and the goal is to predict whether (negative class) or not (positive class) a weapon was discovered on inspection. We use race as a sensitive feature, $z$, in our experiments, with African-Americans ($z=0$) and white ($z=1$) as two values of this feature. After processing the data similar to \citet{zafar_dmt} the dataset consists of $5,832$ subjects and $22$ features.

\subsection{Experimental Setup}\label{sec:exp_setup}
The datasets were split into $50\%$ training, $25\%$ validation and $25\%$ test datapoints. Training data was used to train the classifiers, validation data for tuning hyper parameters and test data to report the results. The CVXPy library \cite{cvxpy} was used to solve all the formulations. We show results using linear classifiers, as decisions made by the linear classifiers are relatively easier to explain, which is an import goal for applications with social significance such as recidivism risk prediction. Additionally, data are likely to be linearly separable in higher dimensions.  
We show some results using nonlinear boundaries with our methods in the appendix. 

\xhdr{Selecting $\CC_{\epsilonb, \thetab_{best}}$} We generate $\CC_{\epsilonb, \thetab_{best}}$ by solving \dscour, given by Problem~\ref{eq:scal_disc_prob}, for a range of $\gamma$ values or \ambour, given by Problem~\ref{eq:scal_amb_prob}, for each training datapoint. Then, we use the validation data to prune the resulting classifiers which lie within a given $\epsilon$ threshold of the most accurate classifier. The results are averaged over 5 runs of these steps using different seed values to initialize the data-split and the solver. For \dscour, we pick the $\CC_{\epsilonb}$ from the aggregated solutions of all the seeds and present the averaged statistics over all the seeds. 
\begin{table}[t]
\caption{[Synthetic dataset] Signed differences in FPR/FNR}
\label{tab:synth_fairness}
\centering
\resizebox{.48\textwidth}{!}{
\begin{tabular}[b]{c||c||c || c ||cc }\hline

\toprule
      &  \multicolumn{3}{c||}  {Unfairness} & Accuracy & \\ 
    \midrule 
                  & total & unamb & amb & \\
    \midrule
    Acc. & -0.13/-0.14 & 0.05/-0.06 & 0.46/-0.45 & 0.89 \\
    Fair & 0.03/-0.02 & 0.05/-0.06 & -0.14/0.29 & 0.77/0.89 \\
    Uni-\ref{eq:scal_disc_prob} & 0.04/-0.04 & 0.05/-0.06 & -0.22/0.20 &  0.89 / 0.89\\
    Our-\ref{eq:scal_disc_prob} & 0.07/-0.07 & 0.05/-0.06 & \bf{0.0}/\bf{-0.01} & 0.89/0.89 \\

    \midrule

\end{tabular}}
\resizebox{0.48\textwidth}{!}{
\begin{tabular}[b]{c||c||c || c ||cc }\hline

     \multicolumn{5}{c}  {With \ref{eq:scal_amb_prob} }  \\ 
            \midrule
    Acc.    & 0.13/-0.14 & 0.06/-0.07 & 0.30/-0.35 & 0.89 \\
    Fair & 0.03/-0.02 & 0.05/-0.07 & -0.06/0.18 & 0.77/0.89 \\
    Uni-\ref{eq:scal_amb_prob} & 0.10/-0.10 & 0.06/-0.07 & 0.16/-0.16 &  0.88 / 0.88\\
    Our-\ref{eq:scal_amb_prob} & 0.06/-0.07 & 0.06/-0.07 & \bf{0.01}/\bf{-0.03} & 0.88/ 0.88 \\

    \midrule

\end{tabular}}
\caption*{
        This table demonstrates that our method is effective in removing unfairness at a very small cost of decrease in the accuracy. Please refer to Section~\ref{sec:eval_synth}
    }
\vspace{-2.5mm}
\end{table}

 We assume that $\epsilon$ is chosen by the experts for the prediction task at hand. We present results for $\epsilon = 0.02$ for the synthetic dataset, and $\epsilon = 0.01$ for real-world datasets. We experimented with several values of $\epsilon$ and obtained similar results. \subsection{Benchmarks and Metrics}

In this section, we discuss the benchmarks and metrics we used to evaluate our proposals. 

\xhdr{Ambiguous regions computation benchmarks} In order to demonstrate the efficiency of our methods to identify the ambiguous regions using \dscour and \ambour, we compare with \dscold and \ambold. We solved the \dscold and \ambold problems using the CPLEX library, with the code provided by the authors~\cite{marx2019predictive}.

\xhdr{Metrics for evaluating ambiguous regions computation} Since the best classifiers for non-scalable and our scalable methods, i.e., $\phib_{best}$ and $\thetab_{best}$, are different, we report the ambiguity $\hat{\alpha}$ and discrepancy $\hat{\delta}$ between any two classifiers in $\CC_\epsilon$, for the respective methods. They are formally defined as follows: 
\begin{equation}\label{eq:non_scal_disc_reported}
 \hat{\delta}_{\epsilonb}(\phib) = \max_{\phib, \hat{\phib} \in \CC_{\epsilonb}} \frac{1}{n} \sum_{\xb_i} \II[\phib(\xb_i) \neq \hat{\phib}(\xb_i)] 
\end{equation}
\begin{equation}\label{eq:non_scal_amb_reported}
  \hat{\alpha}_{\epsilonb}(\phib) =  \frac{1}{n} \sum_{\xb_i} \max_{\phib, \hat{\phib} \in \CC_{\epsilonb}} \II[\phib(\xb_i) \neq \hat{\phib}(\xb_i)].
 \end{equation}
High values of these measures are desired, as that would imply that the $\CC_{\epsilonb}$ contains diverse classifiers which can identify more number of datapoints that have a contradictory decision for a given value of $\epsilonb$. We also report the time it takes to compute the set of classifiers $\CC_{\epsilonb}$. 

\begin{table}[t]
\caption{Comparison identifying ambiguous regions}
\centering
\resizebox{0.48\textwidth}{!}{
\begin{tabular}[b]{c||cc||cc || cc ||cc || cc}\hline

\toprule
     $\epsilonb$ &  \multicolumn{2}{c}{ \ref{eq:non_scal_disc_prob}} & \multicolumn{2}{c}{ \ref{eq:non_scal_amb_prob}} &\multicolumn{2}{c}{ \ref{eq:scal_disc_prob}} & \multicolumn{2}{c}{ \ref{eq:scal_amb_prob}} & \\ 
      \midrule
   -    & $\hat{\delta}$ & $\hat{\alpha}$& $\hat{\delta}$ & $\hat{\alpha}$ & $\hat{\delta}$ & $\hat{\alpha}$& $\hat{\delta}$ & $\hat{\alpha}$ \\
 0.03  & 0.15 & 0.16 & \bf{0.18} & \bf{0.28} & 0.14 & 0.16 & \bf{0.18} & 0.26 \\
 0.05  & 0.17 & 0.19 & 0.22 & \bf{0.38} & 0.16 & 0.17 & \bf{0.23} &  0.36 \\
 0.09  & 0.22 & 0.24 & \bf{0.32} & \bf{0.56}  & 0.2 & 0.20 & \bf{0.32} & 0.51 \\

\end{tabular}}
\centering
\resizebox{0.25\textwidth}{!}{
\begin{tabular}[b]{c||cc||cc||cc||cc||}\hline
    \multicolumn{5}{c} {Training Time}  \\
     $Time$ & \ref{eq:non_scal_disc_prob}  & \ref{eq:non_scal_amb_prob} & \ref{eq:scal_disc_prob}& \ref{eq:scal_amb_prob} \\
    mins & 510  &  19227 &  5 & 5   
\end{tabular}}
\caption*{\label{tab:table_comp} The table above shows maximum discrepancy and ambiguity between any two classifiers in the $\CC_{\epsilonb, \psi : \psi \in \{\phib_{best}, \thetab_{best}\}}$. The bottom table shows the time it took to compute the ambiguous regions with each method. It shows that our methods, given by \ref{eq:scal_disc_prob} and \ref{eq:scal_amb_prob}, achieve comparable performance compared to \ref{eq:non_scal_disc_prob} and \ref{eq:non_scal_amb_prob} and they are upto four orders of magnitude faster. Please refer to Section~\ref{sec:eval_synth}}
\end{table}

\xhdr{Fairness benchmark} For results on fairness in the ambiguous regions, we compare our method given by Problem~\ref{eq:amb_fairness} using $\CC_{\epsilonb,\thetab_{best}}$, picking classifiers uniformly at random from $\CC_{\epsilonb,\thetab_{best}}$, the most accurate classifier and a traditional fair classifier. We chose one traditional fair method as a baseline, as \citet{zafar_fairness} show comparison to other approaches and get similar results. Its formulation (\cite{zafar_fairness} and \cite{ali2019}) is given as follows,
\begin{align}\label{eq:fair_clf}
&\mbox{minimize}\quad - \frac{1}{|\Dcal|} \sum_{(\xb,y)\in\Dcal} \log p(y | \xb, \thetab) + \lambda||\thetab|| \tag{P6}
\\ \nonumber
&\mbox{subject to} \quad \frac{1}{|\Dcal_{*}|}~
\bigg | 
\sum_{(\xb,z) \in \Dcal_{*}}  (z - \bar{z}) d_{\thetab}(\xb_i) 
\bigg| < c, \\ \nonumber 
\end{align}
where $\Dcal_{*}$ was set to datapoints with ground truth negative labels and ground truth positive labels for equalizing false positive rates (FPR) and false negative rates (FNR), respectively. $z$ represents the value of the sensitive feature and $c$ represents the allowed correlation between $z$ and the decision boundary, $d_\theta$. 

We train accurate classifiers by solving  \\$\mbox{minimize}\quad - \frac{1}{|\Dcal|} \sum_{(\xb,y)\in\Dcal} \log p(y | \xb, \thetab) + \lambda||\thetab||$ for different $\lambda$.
Logistic regression loss was used to train all the classifier. More details such as ranges for the hyper parameter search values, seeds, specifications of the machines used and other training details are included in the appendix.

\xhdr{Metrics for fairness} We assume a binary valued sensitive attribute and report a signed difference of FPR and FNR between the unprotected and the protected group for the sensitive feature $z$. \begin{align}\label{eq:fair_metric}
\text{unfairness-FPR} &= FPR_{z=1} - FPR_{z=0}, \\
\text{unfairness-FNR} &= FNR_{z=1} - FNR_{z=0}
\end{align}
We present these numbers for the overall  data, for the unambiguous regions, i.e., where all the classifiers give unanimous decisions, and for the ambiguous regions. We also report the accuracies. We aim to achieve low disparity in group error rates in the ambiguous regions, while achieving an accuracy similar to the most accurate classifier.

\begin{table}[t]
    \caption{[COMPAS] Signed differences in FPR/FNR}
    \label{tab:compas_fairness}
    \centering
    \resizebox{0.48\textwidth}{!}{
        \begin{tabular}[b]{c||c||c || c ||cc }\hline
            
            \toprule
            &  \multicolumn{3}{c||}  {Unfairness} & Accuracy & \\ 
            \midrule 
            & total & unamb & amb & \\
            \midrule
            Acc. & -0.19/0.33 & -0.23/0.41 & \textbf{0.08}/-0.20 & 0.66 \\
            Fair & 0.02/0.03 & -0.09/0.18 & 0.83/-0.92 & 0.66/0.65 \\
            Uni-\ref{eq:scal_disc_prob} & -0.20/0.35 & -0.23/0.41 & \textbf{-0.08}/\textbf{-0.004} &  0.66 /0.66\\
            Our-\ref{eq:scal_disc_prob} & -0.20/0.35 & -0.23/0.41 & \textbf{-0.08}/-0.02 & 0.66/0.66 \\
            
            \midrule

    \end{tabular}}
    
    \resizebox{0.48\textwidth}{!}{
        \begin{tabular}[b]{c||c||c || c ||cc }\hline
            
                        \multicolumn{5}{c}  {With \ref{eq:scal_amb_prob} }  \\ 
                                    \midrule
            Acc.    & -0.19/0.33 & -0.24/0.54 & -0.11/0.15 & 0.66 \\
            Fair & 0.02/0.03 & -0.24/0.54 & 0.34/0.-0.42 & 0.66/0.65 \\
            Uni-\ref{eq:scal_amb_prob} & -0.19/0.34 & -0.24/0.54 & -0.11/0.15 &  0.66/ 0.66\\
            Our-\ref{eq:scal_amb_prob} & -0.14/0.26 & -0.24/0.54 & \textbf{-0.01}/\textbf{0.03} & 0.66/ 0.66 \\
            
            \midrule

    \end{tabular}}
    \caption*{
        This table demonstrates that our methods are effective in removing unfairness in the ambiguous regions at no expense of accuracy. Please refer to Section \ref{sec:eval_real}
    }
\end{table}

\subsection{Synthetic Experiments}\label{sec:eval_synth}
In this section, we answer the evaluations questions using the synthetic dataset. 

\xhdr{Q1: Ambiguous regions coverage and speed} We compared our methods, \dscour and \ambour, of identifying the ambiguous regions with \dscold and \ambold. Table~\ref{tab:table_comp} reports the time it took to compute the ambiguous regions as well as the metrics described in Section~\ref{sec:exp_setup}. The results demonstrates that our methods are comparable or even better in coverage of the ambiguous regions on the test data, while being up to four orders of magnitude faster. 

\xhdr{Q2: Accuracy fairness trade-off} We compare our method with the benchmarks described in Section~\ref{sec:exp_setup}. The results in Table \ref{tab:synth_fairness} demonstrate that:

Existing fairness methods sometime achieves overall fairness at the expense of a significant decrease in accuracy. Additionally, overall fairness is achieved by being biased towards different groups for different types of errors, i.e., ones in the unambiguous vs ambiguous regions. On the other hand, our method is effective in removing unfairness in the ambiguous regions and ignoring the unfairness in the unambiguous regions, as desired. Our method also achieves accuracy similar to the most accurate classifiers.

\xhdr{Q3: Robustness to noisy data} In order to demonstrate the sensitivity of existing fairness methods towards noise, we flipped the ground truth labels of $0.0\%$ to $20\%$ of the datapoints uniformly at random. Figures~\ref{fig:clf_fair} and~\ref{fig:clf_ours} present our findings. 
We compare an accurate classifier, a fair classifier and our method equalizing FPR using \ambour. The key takeaways are as follows: In an effort to equalize all errors, existing fairness methods are affected by label noise and end up classifying a significant number of datapoints in the wrong class, as hypothesized in the introduction. In contrast, our method is robust to noise as it identifies similar regions as ambiguous for varying level of noise. Secondly, this experiment also 
confirms our hypothesis, by showing that the ambiguous region coincide with regions with predominantly high model uncertainty, i.e., the sparse clusters. 

\subsection{Evaluation on Real-World Datasets}\label{sec:eval_real}

In this section, we answer our evaluation questions using two real-world datasets.

\xhdr{Q1: Ambiguous regions coverage and speed} We identify the datapoints with ambiguous decisions using \dscour, given by Problem~\ref{eq:scal_disc_prob} and \ambour, given by Problem~\ref{eq:scal_amb_prob}, for the same value of $\epsilonb$. We also tried \dscold and \ambold, however after several hours of computations they still did not yield any results. So, we compare the results of our two proposals, using $\hat{\alpha}$ metric given by Equation~\ref{eq:non_scal_amb_reported}. Takeaways remain similar for $\hat{\delta}$, given by Equation~\ref{eq:non_scal_disc_reported}. 

For the Compas data, our method \dscour and \ambour categorized $0.12$ and $0.5$ of the datapoints as having an ambiguous decision, respectively. While for the SQF dataset, $0.12$ and $0.53$ of the datapoints were identified as having an ambiguous decision by \dscour and \ambour, respectively. It is noteworthy that \ambour identifies more datapoints as ambiguous. This is due to the fact that with \ambour we train one classifier per training datapoint, i.e., we perform a more exhaustive search for the classifiers that exhibit predictive multiplicity. This process, however, takes a longer time. Hence, there is a trade-off between the speed and effectiveness for both the proposed methods of identifying the ambiguous regions.

\xhdr{Q2: Accuracy fairness trade-off}
Similar to the synthetic dataset, we compare our method of equalizing group error rates (FPR and FNR) in the ambiguous regions, identified by \dscour and \ambour, with three benchmarks described in Section~\ref{sec:exp_setup}. The takeaways from results presented in Tables~\ref{tab:compas_fairness} and \ref{tab:sqf_fairness} are the following. 
Existing fair classifiers that focus on equalizing overall error have high unfairness in the ambiguous regions in most cases, which confirms our hypothesis. Although these classifiers achieve fairness in the overall data, they sometimes result in a significant drop in accuracy. Additionally, in many cases, existing fair classifiers achieve overall fairness by being unfair to different groups in the ambiguous vs unambiguous regions.

In comparison, our method that only equalizes errors in the ambiguous regions, in most cases, provides the fairest solution in the ambiguous regions while achieving a comparable accuracy to the most accurate classifier.

In a few cases where our approach is not the only best solution, it provides additional benefits, e.g., in one case our solution is equally fair in the ambiguous region compared to the accurate classifier (cf. Table~\ref{tab:sqf_fairness}). However, our method assigns decisions to datapoints in the ambiguous regions stochastically. So, in practice, most datapoint in the ambiguous region have a non-zero probability to be in the favorable class. 
This is desirable over a deterministic decision, since there is ambiguity in decisions for these datapoints. In another case, Table~\ref{tab:compas_fairness}, selecting classifiers uniformly at random is $1.6\%$ more fair on the \textit{test data}. However, our solution is still $90\%$ and $18\%$ better than the benchmark fair classifier and the accurate classifier, which are the current standards.

\begin{table}[t]
    \caption{[SQF] Signed differences in FPR/FNR}
    \label{tab:sqf_fairness}
    \centering
    \resizebox{0.48\textwidth}{!}{
        \begin{tabular}[b]{c||c||c || c ||cc }\hline
            
            \toprule
            &  \multicolumn{3}{c||}  {Unfairness} & Accuracy & \\ 
            \midrule 
            & total & unamb & amb & \\
            \midrule
            Acc. &-0.28/0.12 & -0.29/0.13 & -0.07/\textbf{0.017} & 0.75 \\
            Fair & 0.04/0.02 & 0.02/0.03 & 0.07/-0.15 & 0.65/0.71 \\
            Uni-\ref{eq:scal_disc_prob} & -0.28/0.12 & -0.29/0.13 & -0.05/\textbf{0.014} &  0.75 / 0.75\\
            Our-\ref{eq:scal_disc_prob} &-0.28/0.11 & -0.29/0.13 & \textbf{-0.02}/\textbf{-0.017} & 0.75/ 0.75 \\
            \midrule    
    \end{tabular}}
    
    \resizebox{0.48\textwidth}{!}{
        \begin{tabular}[b]{c||c||c || c ||cc }\hline
            
                        \multicolumn{5}{c}  {With \ref{eq:scal_amb_prob} }  \\ 
                                    \midrule
            Acc.   & -0.28/0.12 & -0.24/0.17 & -0.25/\textbf{0.07} & 0.75 \\
            Fair & 0.04/0.02 & -0.06/0.12 & \textbf{0.15}/-0.08 & 0.65/0.71 \\
            Uni-\ref{eq:scal_amb_prob} & -0.27/0.14 & -0.24/0.17 & -0.25/0.09 &  0.74/ 0.74\\
            Our-\ref{eq:scal_amb_prob} & -0.24/0.13 & -0.24/0.17 & {-0.18}/\textbf{0.07} & 0.73/ 0.74 \\            
            \midrule            
            
    \end{tabular}}
    
    \caption*{This table demonstrates effectiveness of our methods. Please refer to Section \ref{sec:eval_real}}
\end{table}

\section{Related Work}

\xhdr{Fairness in ML}
In recent years a number of fairness methods and notions have been proposed for classification tasks ~\cite{zafar_dmt,Dwork2012,hardt_nips16,goel_cost_fairness,pedreschi_discrimination,zafar_preferred,zafar_fairness,feldman_kdd15,icml2013_zemel13,ali2019,lahoti2019operationalizing,grgic2018beyond, speicher2018unified,lahoti_neurips}. A family of these methods aim to enforce fairness across socially salient groups in the society that equalize 'total' errors e.g., \cite{zafar_dmt,hardt_nips16,ali2019,zafar_preferred}. In contrast, we argue to focus only on the errors arising due to \textit{model uncertainty}. We do so by building on existing work in predictive multiplicity. \\

\xhdr{Modeling uncertainty} Prior works on categorizing uncertainties have proposed to distinguish between aleatoric (irreducible) uncertainty and model (reducible) uncertainty\cite{hora1996aleatory,der2009aleatory, hullermeier2021aleatoric}. A lot of works in machine learning have addressed this distinction in different subfields. \citet{depeweg2018decomposition} propose to decompose the two types of uncertainties using bayesian neural networks and latent variables. \citet{kendall2017uncertainties} consider this distinction in computer vision problems. \citet{mcallister2017bayesian} distinguish between the types of uncertainties in reinforcement learning problems. 

We believe that we are the first ones to propose to distinguish between different types of uncertainties for fairness in predictive tasks. \\

\xhdr{Predictive multiplicity} In their seminal work, \citet{breiman2001statistical} introduced the concept of the \emph{Rashomon effect} in the context of model explanations. The Rashomon effect refers to the scenario where data admits multiple different models that yield similar accuracy. \citet{breiman2001statistical} argue that one should not use the explanations of a single model to draw conclusions about the data and the prediction task at hand. 
Rashomon sets, defined as $\epsilon$-set of models, i.e.  those whose empirical training loss is within $\epsilon$-loss of a baseline classifier, are used by 
\citet{fisher2019all,dong2019variable} to study the problem of variable importance. 

The notion of predictive multiplicity in a classification setting was introduced by \citet{marx2019predictive}. 
They proposed mixed integer programming methods using non-convex loss functions to train classifiers which would yield predictive multiplicity for linear classifiers. 
We build on this work, and extend it by proposing tractable convex problem formulations which yield fast solutions, and work for both linear and non-linear classifiers.

There is a growing interest in predictive multiplicity due to its societal implications on algorithmic decision-making system. \citet{bhattcounterfactual} look at it from a fairness perspective, and aim to find counterfactual accuracy of a classifier which would give a selected test datapoint favorable outcome. Specifically, they aim to find the minimum decrease in accuarcy, $\epsilonb$, that would give an individual a favorable outcome. \citet{pawelczyk2020counterfactual} provide an upper bound for the costs of finding counterfactual explanations under predictive multiplicity. However, none of these works have made the connections between predictive multiplicity and model uncertainty.

\section{Concluding discussion}

In this work, we propose that while designing fairness approaches one must account for the uncertainties of the prediction task at hand. Specifically, we argue that only the \emph{errors} arising due to lack of knowledge about the best model or due to lack of data, i.e., the \emph{epistemic errors} should be taken into account while designing fairness methods and errors due to inherent noise should be ignored. 
Our proposal stands in contrast to the current group fairness approach that aims to equalize 'total' errors. With this goal in mind, we build upon predictive multiplicity techniques to identify the regions with model uncertainty.

In addition, we propose convex and scalable formulations to find classifiers that exhibit predictive multiplicity, which are approximately equally effective 
compared to their non-convex counterparts, while being up to four orders of magnitude faster. We also propose convex formulations to equalize errors arising due to model uncertainty. Using synthetic and real-world datasets, we demonstrate that our methods are effective and more robust to label noise compared to existing group fairness methods.

Our key insight is that not all types of errors are equal and when improving parity in the error rates one must account for the type of uncertainty inducing the error. We believe that this insight and our predictive multiplicity methods open new avenues for research on how to account for uncertainties when designing fair machine learning methods.

\section*{Acknowledgements}
This research was supported in part by a European Research Council (ERC) Advanced Grant for the project “Foundations for Fair Social Computing”, funded under the European Union’s Horizon 2020 Framework Programme (grant agreement no. 789373). This work was also partly supported by the ERC Synergy Grant 610150 (imPACT).

\clearpage
\bibliographystyle{ACM-Reference-Format}
\balance
\bibliography{multiplicity_fairness}

\clearpage
\begin{appendix}

\begin{figure*}[ht]
 \centering
      \subfloat         {
     \includegraphics[angle=0, width=0.55\columnwidth]{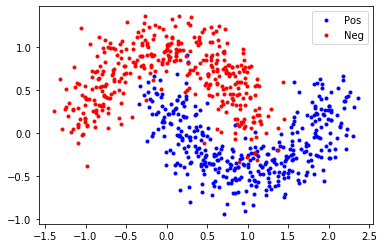}
         } 
     \subfloat     {
     \includegraphics[angle=0, width=0.55\columnwidth]{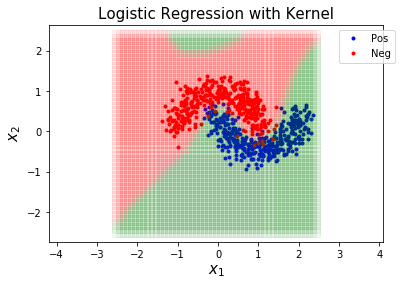}
         }
     \subfloat     {
         \includegraphics[angle=0, width=0.55\columnwidth]{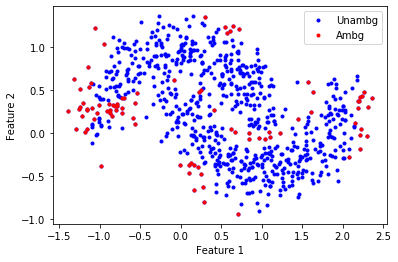}
         }

     \caption{
    [Synthetic dataset-non-linear] The figure on the left shows the 2 moons dataset, the middle figure shows the best non-linear boundary with green regions classified as positive and red regions as negative and the one on the right shows the ambiguous regions identified using our method. The figure demonstrate that unlike ~\citet{marx2019predictive} our methods can also be used to identify predictive multiplicity for non-linear classifiers.
    }
    \label{fig:non_lin}
 \end{figure*}
  \begin{figure*}[!ht]
 \centering
      \subfloat         {
     \includegraphics[angle=0, width=0.50\columnwidth]{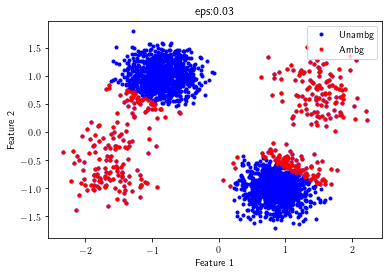}
         }
     \subfloat     {
     \includegraphics[angle=0, width=0.50\columnwidth]{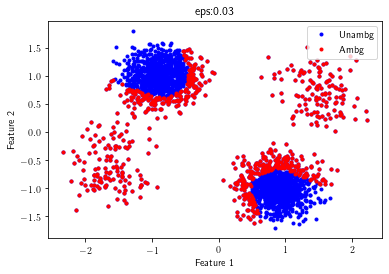}
         }
     \subfloat     {
         \includegraphics[angle=0, width=0.45\columnwidth]{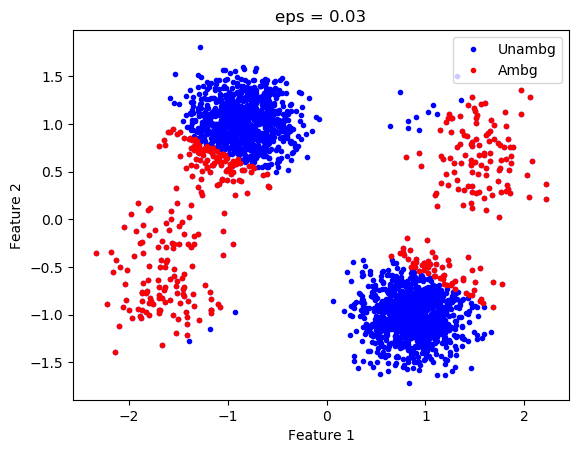}
         }
      \subfloat     {
         \includegraphics[angle=0, width=0.45\columnwidth]{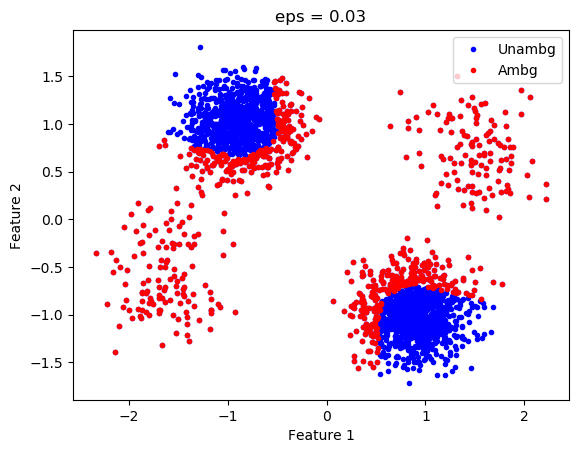}
         }

     \caption{
    [Synthetic dataset] This figure shows the ambiguous regions identified by the four methods discussed in the paper. From left to right figures corresponds to \dscold, \ambold, \dscour, \ambour. It demonstrates that our methods identify similar ambiguous regions compared to the exact methods proposed by ~\citet{marx2019predictive}. 
    }
    \label{fig:ambg_regions}
 \end{figure*}
\section{Training details}

In this section we explain the training details for our methods.

In order to train \dscour and \ambour, presented in Section 4.1 of the paper,  we used CPLEX library~\cite{boyd_concave_convex}. \dscour is give as follows,
\begin{align}\label{eq:scal_disc_prob}
&\underbrace{\min_{\thetab} - \frac{1}{N} \sum_{\xb_i,y_i} p(y_i|\xb_i; \thetab)}_\text{maximize accuracy} \tag{P3}\\
&\text{subject to: } \underbrace{\frac{1}{N} \sum_{\xb_i} \max(0, d_{\thetab(\xb_i)} d_{\thetab_{best}(\xb_i)}) \leq \gamma}_\text{limit agreement to $\thetab_{best}$} \nonumber
\end{align}

For synthetic dataset described in the paper we trained $1000$ classifiers with $\gamma \in (1e-15 , 2.0)$ picked linearly. For SQF dataset we also trained $1000$ classifiers with $\gamma \in (0.0 , 2.0)$  and for compas dataset we trained $1000$ classifiers with $\gamma \in (0.0 , 10.0)$ picked linearly. 

In order to train the baselines mentioned in the experiment section of the paper, we trained $100$ classifiers using logistic regression with L2 regularizer, $\mbox{minimize}\quad - \frac{1}{|\Dcal|} \sum_{(\xb,y)\in\Dcal} \log p(y | \xb, \thetab) + \lambda||\thetab||$ , with $\lambda \in (1e-1, 1)$, where $p(y = 1 | \xb, \thetab) = \frac{1}{1 + \exp(-\thetab^T\xb)}$. We picked the $\lambda$ that yielded the best accuracy on the validation set. 

For traditional fairness methods given by, 

\begin{align}\label{eq:fair_clf}
&\mbox{minimize}\quad - \frac{1}{|\Dcal|} \sum_{(\xb,y)\in\Dcal} \log p(y | \xb, \thetab) + \lambda||\thetab|| \tag{P6}
\\ \nonumber
&\mbox{subject to} \quad \frac{1}{|\Dcal_{*}|}~
\bigg | 
\sum_{(\xb,z) \in \Dcal_{*}}  (z - \bar{z}) d_{\thetab}(\xb_i) 
\bigg| < c, \\ \nonumber 
\end{align}

where $p(y = 1 | \xb, \thetab) = \frac{1}{1 + \exp(-\thetab^T\xb)}$ and $z$ is the sensitive attribute, same $\lambda$ was used which we picked by training the accurate classifier. We trained $100$ fair classifiers for each dataset by varying $c$ values, which could be written as the product of correlation between different the sensitive attribute and $\thetab_{best}$ and multiplicative factor varying between zero and 1 ~\cite{zafar_fairness}, i.e., $c = t \cdot cov(\thetab_{best},z)$. For synthetic dataset we used we use $t \in (0,0.2)$ and for real world datasets $t \in (0.0 , 1e-5)$. We train a pool of benchmark fair classifiers for varying values of $c$ and a pool of accurate classifiers on 5 different shuffles of the data and then pick the fairest classifier and most accurate classifiers, respectively, for each shuffle from this pool. 

We aggregated the results using these  $5$ seed values, $[1122334455, \\2211334455, 1133224455, 3322441155, 1122443355]$. We used Intel(R) Xeon(R) CPU E7-8857 v2 @ 3.00GH with 48 cores to run all the experiments. 

\section{Predictive multiplicity comparison}
In this section we show the visualization of the ambiguous regions with different methods introduced in the paper. Figure~\ref{fig:ambg_regions} shows ambiguous regions identified by the exact methods proposed by \citet{marx2019predictive}, \dscold and \ambold, and our methods \dscour and \ambour. The figure demonstrates that visually our methods identify similar regions with ambiguous results. In general, we also see that ambiguous regions are the more sparse regions of feature space, where decisions are difficult to make.

\subsection{Results using nonlinear Classifiers}
In this section we show the results using kernalized logistic regression to identify ambiguous regions, with \dscour. Figure~\ref{fig:non_lin} demonstrate the results.

\end{appendix}
\end{document}